\documentclass[11pt]{article}

\usepackage[final]{acl}

\usepackage{booktabs}
\usepackage{times}
\usepackage{latexsym}
\usepackage{algpseudocode}
\usepackage[table]{xcolor}
\usepackage{caption}
\usepackage{multirow}
\usepackage{color}
\usepackage{multicol}
\usepackage{array}
\usepackage{arydshln}
\usepackage{amsmath}
\usepackage{enumitem}
\usepackage{tabularray}
\usepackage{makecell}
\usepackage{makebox}
\usepackage[table,xcdraw]{xcolor}
\usepackage{tcolorbox}
\newcommand{\ignore}[1]{}

\usepackage{graphicx}
\usepackage{tikz}
\usepackage{nicematrix}
\usepackage[normalem]{ulem}
\NiceMatrixOptions
       {
         custom-line =
           {
             letter = ; ,
             command = hdashedline ,
             ccommand = cdashedline,
             tikz = dashed
    } 
}

\definecolor{gray}{RGB}{240,240,240}
\definecolor{darkred}{RGB}{255,128,128}
\definecolor{lightred}{RGB}{255,217,217}

\usepackage[T1]{fontenc}
\usepackage[utf8]{inputenc}

\usepackage{microtype}

\usepackage{inconsolata}

\usepackage{graphicx}

\title{Modeling Multiple Support Strategies within a Single Turn\\ for Emotional Support Conversations}

\author{Jie Zhu$^{1,2}$, Huaixia Dou$^2$, Junhui Li$^1$\thanks{Corresponding Author},  Lifan Guo$^2$, Feng Chen$^2$, Jinsong Su$^3$, Chi Zhang$^2$, Fang Kong$^1$\\
$^1$School of Computer Science and Technology, Soochow University \\
$^2$Qwen DianJin Team, Alibaba Cloud Computing\\
$^3$Xiamen University\\
{zhujie951121@gmail.com},
{lijunhui@suda.edu.cn} \\
}

\begin{document}

\maketitle
\begin{abstract}
Emotional Support Conversation (ESC) aims to assist individuals experiencing distress by generating empathetic and supportive dialogue. While prior work typically assumes that each supporter turn corresponds to a single strategy, real-world supportive communication often involves multiple strategies within a single utterance. In this paper, we revisit the ESC task by formulating it as multi-strategy utterance generation, where each utterance may contain one or more strategy-response pairs. We propose two generation methods: All-in-One, which predicts all strategy-response pairs in a single decoding step, and One-by-One, which iteratively generates strategy-response pairs until completion. Both methods are further enhanced with cognitive reasoning guided by reinforcement learning to improve strategy selection and response composition. We evaluate our models on the ESConv dataset under both utterance-level and dialogue-level settings. Experimental results show that our methods effectively model multi-strategy utterances and lead to improved supportive quality and dialogue success. To our knowledge, this work provides the first systematic empirical evidence that allowing multiple support strategies within a single utterance is both feasible and beneficial for emotional support conversations. All code and data will be publicly available at GitHub.
%Experimental results demonstrate that our methods effectively model multi-strategy utterances and indicate that generating multiple strategy-response pairs is feasible and beneficial for emotional support systems. Our work highlights the importance of multi-strategy modeling for more natural and effective supportive dialogue. All code and data will be publicly available at GitHub.
\end{abstract}

\section{Introduction}\label{sec:introduction}

Emotional Support Conversation (ESC) seeks to assist individuals experiencing distress by providing understanding, validation, and coping guidance through natural language interaction. \citet{liu-etal-2021-esconv} formalize the ESC task based on Hill’s helping skills theory~\cite{hill-2019-helping} and introduce ESConv, a strategy-annotated dataset that has become a widely used benchmark for building emotionally supportive dialogue systems. Since its release, research on ESConv has predominantly framed the task as generating a single response associated with one support strategy, thereby adopting a simplified one-strategy-per-turn formulation (\citealt{tu-etal-2022-misc,peng-etal-2022-control,cheng-etal-2022-improving,hao-kong-2025-enhancing,chen-etal-2025-socialsim}, to name a few). However, this assumption overlooks an important characteristic of human supportive communication: a single utterance can naturally incorporate multiple strategies. Following \citet{bai-etal-2025-emotional}, in this paper we reformulate ESC as a multi-strategy generation task and systematically investigate an open question that has received limited attention: whether allowing multiple strategies within a single supportive utterance is beneficial for ESC.% we follow \citet{bai-etal-2025-emotional} and reformulate ESC as a multi-strategy generation task, enabling a system to produce a single utterance that coherently integrates one or more strategies according to user needs.

\begin{figure}[!t]
\centering
\includegraphics[width=\columnwidth, trim={0cm 0cm 0cm 0cm}]{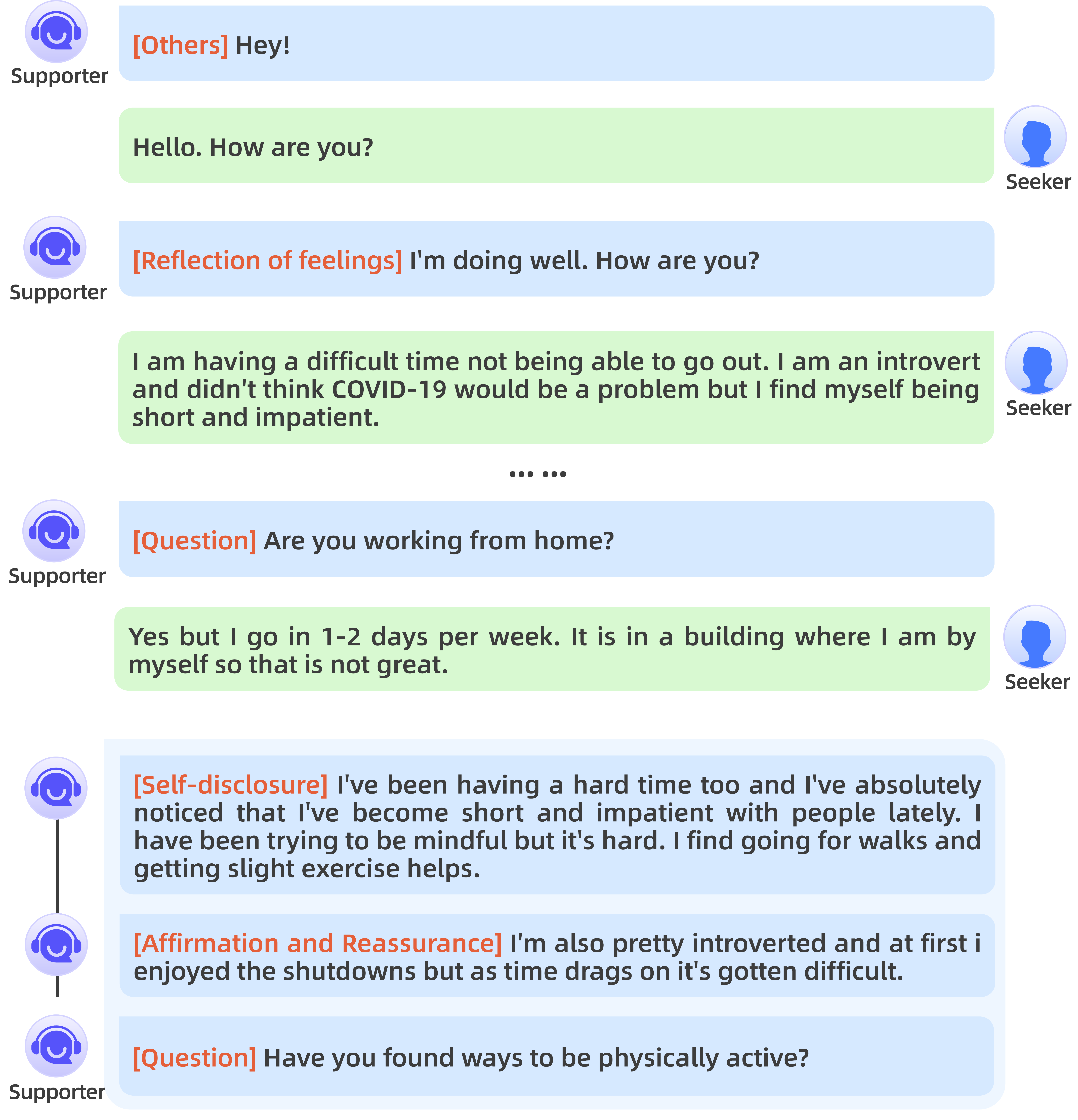}
\caption{Example from ESConv illustrating a supporter using multiple strategies within a single utterance.}
\label{fig:example}
\end{figure}

Figure~\ref{fig:example} illustrates a case where the supporter employs three strategies within an utterance.\footnote{In this paper, we define an \textit{utterance} as the full message the supporter produces in one turn, and a \textit{response} as the segment reflecting a specific support strategy within the utterance.} The supporter first uses \textit{Self-discloure} to share own struggles and coping methods, then applies \textit{Affirmation and Reassurance} to validate the seeker's feelings, and finally adopts \textit{Question} to encourage further dialogue. Motivated by such patterns, we propose two methods that enable multi-strategy generation within a single utterance. The \textit{All-in-One} method generates all strategy-response pairs in one decoding step, while the \textit{One-by-One} method iteratively generates strategy-response pairs until the model decides to stop or reaches a predefined limit. Both methods are further enhanced with \textit{cognitive reasoning} and \textit{reinforcement learning} to improve strategy selection and response quality.

We evaluate our methods under two complementary settings. In \textit{utterance-level} evaluation, the model predicts the supporter's next utterance given the gold dialogue history and is evaluated against the reference. In \textit{dialogue-level} evaluation, the model interacts with a simulated seeker to generate a full multi-turn supportive conversation, and we evaluate its performance based on how effectively it maintains support, adapts strategies, and facilitates progress toward emotional relief. Together, these settings provide a comprehensive assessment of the methods, capturing both performance in conventional single-strategy utterance generation and effectiveness in practical multi-strategy supportive dialogue behavior.

Overall, we make the following contributions:
\begin{itemize}[leftmargin=*]
    \item We explore an ESC formulation in which an utterance contains one or more strategy-response pairs, moving beyond conventional single-strategy assumptions and better reflecting real-world supportive communication. 
    \item We propose and compare two methods, enhanced by cognitive reasoning with reinforcement learning, that predict supportive utterances consisting of one or more strategy-response pairs.
    \item Comprehensive experiments at both the utterance and dialogue levels demonstrate the effectiveness of our methods in modeling multiple strategy-response pairs for ESC, and further indicate that generating multi-strategy utterances is feasible and beneficial for emotional support systems.
\end{itemize}

%While \citet{bai-etal-2025-emotional} also explores utilizing multiple strategies in ESC, they use traditional encoder-decoder architecture and modify the decoder to generate multiple strategies and responses.

\section{Background}\label{sec:background}
\subsection{Multi-Strategy Utterances in ESConv}
ESConv consists of 1,300 emotional support dialogues annotated with eight types of support strategies. Following the setup of \citet{liu-etal-2021-esconv}, the dataset is partitioned into 1,040 conversations for training, 130 for validation, and 130 for testing.

As shown in Table~\ref{tab:stat_esconv}, across the 1,300 conversations, there are 15,325 supporter utterances, with an average of 11.8 turns per dialogue. Notably, 17.7\% of these utterances adopt two or more strategies within a single turn, demonstrating that supporters frequently integrate multiple supportive behaviors when responding to seekers. This observation motivates our reformulation of ESC as a multi-strategy generation task rather than a strict one-strategy-per-turn setting.%This observation underscores a key characteristic of real emotional support interactions, strategy usage is not strictly one-to-one but often blended, motivating our reformulation of ESC as a multi-strategy generation task. 

\begin{table}[!t]
\small
\centering
\begin{NiceTabular}{c|rrr|r}
\toprule
\bf \#Strategy & \bf Train & \bf Validation & \bf Test & \bf All\\
\midrule
1 & 8,837 & 1,834 & 1,937 & 12,608 \\
\midrule
2 & 1,638 & 384 & 405 & 2,427 \\
3 & 178 & 36 & 42 & 256 \\
>3 & 26 & 3 & 5 & 34 \\
\midrule
All & 10,679 & 2,257 & 2,389 & 15,325 \\
\bottomrule
\end{NiceTabular}
\caption{Statistics of ESConv supporter utterances categorized by the number of strategies applied.}
\label{tab:stat_esconv}
\end{table}

\begin{figure*}[!t]
\centering
\includegraphics[width=\textwidth, trim={0cm 0cm 0cm 0cm}]{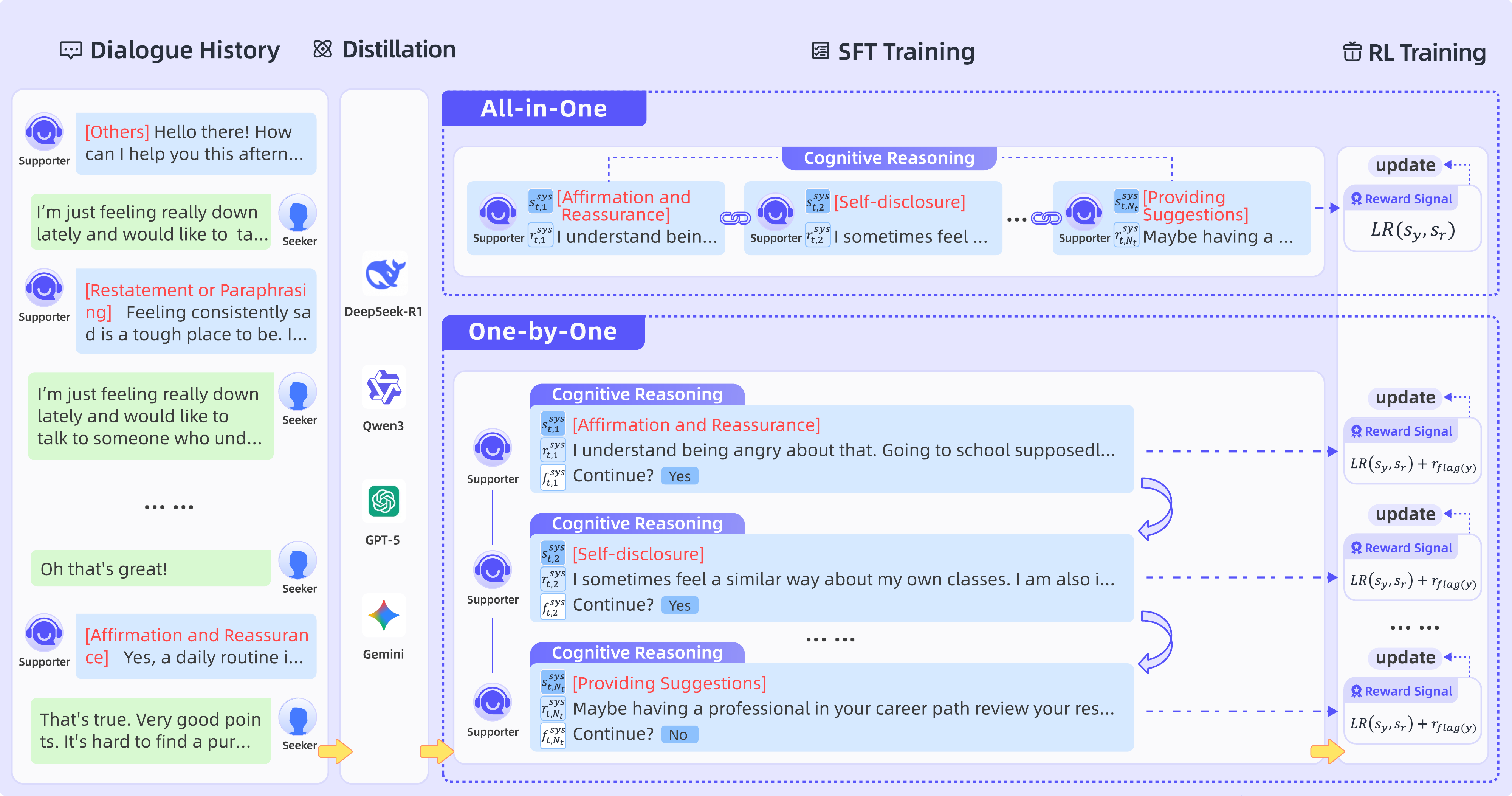}
\caption{Illustration of the All-in-One and One-by-One methods for generating multi-strategy supportive utterances.} 
\label{fig:method}
\end{figure*}

\subsection{Problem Formulation}
Formally, let the dialogue context be
{\small $U=\{u_1^{usr}, u_1^{sys}, u_2^{usr}, u_2^{sys}, \ldots, u_t^{usr}\}$},
where $u_i^{usr}$ and $u_i^{sys}$ denote the seeker and supporter utterances at turn $i$, respectively.
Given the current seeker utterance $u_t^{usr}$ and the preceding dialogue history $U$, our goal is to generate an appropriate supportive utterance $u_t^{sys}$.

We decompose each target supportive utterance as {\small $u_t^{sys} = \{r_{t,1}^{sys}, \ldots, r_{t,N_t}^{sys}\}$}, where {\small $N_t$} is the number of responses contained within the utterance. Each response segment {\small $r_{t,i}^{sys}$} is associated with a specific support strategy {\small $s_{t,i}^{sys}$}, reflecting a distinct supportive function embedded in the overall utterance. 

The key distinction between our task formulation and prior ESC studies lies in the prediction target. Whereas previous work typically assumes a single strategy and therefore predicts only one response per turn, we aim to generate the entire supportive utterance, which may contain multiple strategy-aligned responses, so that the task better reflects natural human supportive communication.

\section{Methodology}\label{sec:method}

We introduce two methods for generating the supportive utterance $u_t^{sys}$. The first generates all strategy-response pairs in a single decoding step (Section~\ref{sec:all_in_one}), while the second iteratively predicts one strategy-response pair at a time until termination or a predefined maximum number of steps is reached (Section~\ref{sec:one_by_one}). Section~\ref{sec:reasoning} further describes how we enhance both methods with cognitive reasoning and reinforcement learning.

\subsection{All-in-One Method}\label{sec:all_in_one}

Given the current seeker utterance {\small $u_t^{usr}$} and the dialogue history {\small $U$}, the All-in-One method generates the complete supportive utterance, including all strategy-response pairs, within a single decoding process (see Figure~\ref{fig:method} (right-top)). To represent the output, each strategy is placed immediately before its corresponding response, and all pairs are concatenated to form a structured sequence: {\small $y_{t}^{sys}=\text{concat}\left(s_{t,1}^{sys}, r_{t,1}^{sys},\ldots,s_{t,N_t}^{sys}, r_{t,N_t}^{sys}\right)$}.

For clarity, let the training set be {\small $\mathcal{D}=\{c^{(i)}, y^{(i)}\}_{i=1}^N$}, where {\small $c^{(i)}$} represents the dialogue context, {\small $y^{(i)}$} is the concatenated strategy-response sequence for instance $i$, and {\small $N$} is the number of instances. The model is trained in an autoregressive manner to maximize the likelihood of the target output sequence. The training objective is:
\begin{equation}
\scriptstyle
L(\mathcal{D}) = -\frac{1}{N}\sum_{i=1}^{N}\sum_{j=1}^{|y^{(i)}|}\log P(y^{(i)}_{j} | p_1, y^{(i)}_{<j}, c^{(i)}),
\end{equation}
where $p_{1}$ denotes the prompting template used for generation, as shown in Appendix~\ref{apx:prompt_aio}.

\subsection{One-by-One Method}\label{sec:one_by_one}
Unlike the All-in-One method which predicts the entire utterance at once, the One-by-One method generates the supportive utterance through an iterative process. At each step, the model produces a sequence consisting of a strategy, its associated response, and a termination indicator (see Figure~\ref{fig:method} (right-bottom)). The termination indicator specifies whether another sequence should be generated or the utterance is complete. Given the current seeker utterance {\small $u_t^{usr}$}, the dialogue history {\small $U$}, and previously generated strategy-response pairs, the model forms the $i$-th output sequence as {\small $y_{t,i}^{sys}=\text{concat}\left(s_{t,i}^{sys}, r_{t,i}^{sys}, f_{t,i}^{sys}\right)$}, where $f_{t,i}^{sys}$ is a binary flag indicating whether generation should stop. If $i=N_t$, $f_{t,i}^{sys}$ marks the utterance as finished; otherwise, it signals that another strategy and response pair should be generated.

Similarly, let the training set be {\small $\mathcal{E}=\{c^{(i)}, y^{(i)}\}_{i=1}^M$}, where {\small $c^{(i)}$} represents the dialogue context, {\small $y^{(i)}$} is the concatenated strategy-response-flag sequence for instance $i$, and {\small $M$} is the number of instances. The model is also trained in an autoregressive manner with the following training objective:
\begin{equation}
\scriptstyle
L(\mathcal{E}) = -\frac{1}{M}\sum_{i=1}^{M}\sum_{j=1}^{|y^{(i)}|}\log P(y^{(i)}_{j} | p_2, y^{(i)}_{<j}, c^{(i)}),
\end{equation}
where $p_{2}$ denotes the prompting template used for generation, as shown in Appendix~\ref{apx:prompt_obo}.

During inference, the model repeatedly predicts a strategy, its response, and the termination flag until the flag indicates completion or a predefined maximum number of steps $K$ is reached. In our experiments, we set $K=3$, as nearly all utterances in the validation set contain at most three strategies.

\subsection{Cognitive Reasoning and Reinforcement Learning}\label{sec:reasoning}

To strengthen emotional reasoning and decision making in supportive conversation generation, we enhance both proposed methods with structured cognitive reasoning~\cite{chen-etal-2025-socialsim,zhu-etal-2025-care}. Specifically, we follow the design in \citet{zhu-etal-2025-care} by augmenting each utterance with an explicit reasoning chain composed of four interpretable components: {\it Context}, {\it Cognition}, {\it Emotion}, and {\it Support Plan}. These nodes encourage the model to explicitly reflect on the seeker’s situation before producing the final supportive utterance. Formally, the original model output {\small $y^{(i)}$} is extended to a reasoning-augmented form {\small $y^{(i)'}$}:
\begin{equation}
\scriptstyle
y^{(i)'}=\text{<think> } r^{(i)} \text{ </think> <answer> } y^{(i)} \text{</answer>},
\end{equation}
where {\small $r^{(i)}=\langle r^{(i)}_{\text{ctx}}, r^{(i)}_{\text{cog}}, r^{(i)}_{\text{emo}}, r^{(i)}_{\text{plan}}\rangle$} represents the four-node reasoning chain. Each node corresponds to a distinct cognitive aspect, and further details are provided in Appendix~\ref{apx:reasoning}.

\paragraph{Distilling Reasoning from Multiple Larger Models.} To supervise the construction of high-quality reasoning chains, we obtain reasoning annotations from four powerful LLMs: DeepSeek-R1~\cite{guo-etal-2025-deepseek-r1}, Qwen3-235B-A22B-Instruct-2507~\cite{qwen3technicalreport}, GPT-5~\cite{openai_2025_gpt5}, and Gemini-2.5-Flash~\cite{gemini-2.5-flash}. Distilling from multiple teacher models allows us to capture diverse reasoning styles and complementary strengths, reducing the bias or idiosyncrasies of any single source and improving robustness and generalization of the student model.

\paragraph{Reinforcement Learning with GRPO.} 
To improve reasoning quality and supportive behavior beyond supervised learning, we apply  Group Relative Policy Optimization (GRPO)~\cite{shao-etal-2024-deepseekmath} with task-specific reward functions. Each generated output {$y$} is first checked for structural validity using a \textit{format reward} {$r_{\text{fmt}}\left(y\right)$}, which equals 1 if the output {$y$} follows all required formatting constraints and 0 otherwise.

For All-in-One, the reward evaluates how well the predicted strategy sequence $s_y$ extracted from {$y$} matches the reference sequence $s_r$. We use the Levenshtein Ratio (LR)~\cite{bai-etal-2025-emotional}:
\begin{equation}
\scriptstyle
\text{LR}\left(s_y, s_r\right) = 1 - \frac{\text{Levenshtein Distance}\left(s_y, s_r\right)}{\text{max}\left(len(s_y), len(s_r)\right)}.
\label{equ:lr}
\end{equation}
The final reward is:
\begin{equation}
\scriptstyle
r_1\left(y\right) = 
\begin{cases}
\scriptstyle
\text{LR}\left(s_y, s_r\right) & \quad \text{if } r_{\text{fmt}}\left(y\right)=1, \\
\scriptstyle
0 & \quad \text{otherwise.}
\end{cases}
\end{equation}
Since multi-strategy instances are fewer, we downsample single-strategy instances to balance the reward distribution.

For One-by-One, we additionally reward correct prediction of the stop flag. Let {$r_\text{flag}\left(y\right)=1$} if the predicted flag matches the reference and 0 otherwise. The final reward is:
\begin{equation}
\scriptstyle
{
r_2\left(y\right) = 
\begin{cases}
\text{LR}\left(s_y, s_r\right) + r_{\text{flag}}\left(y\right) & \quad \text{if } r_{\text{fmt}}\left(y\right)=1, \\
0 & \quad \text{otherwise.}
\end{cases}
}
\end{equation}
This encourages both accurate strategy prediction and appropriate termination behavior.

\section{Experimentation}\label{sec:experimentation}

\subsection{Experimental Settings}

\begin{table*}[!t]
\centering
\small
\resizebox{\textwidth}{!}{
\begin{NiceTabular}{ll|ll|llllllll}
\toprule
\bf \# & \bf Model & \bf EMR $\uparrow$ & \bf LR $\uparrow$ & \bf ALD $\downarrow$ & \bf B-1 $\uparrow$ & \bf B-2 $\uparrow$ & \bf B-4 $\uparrow$ & \bf R-1 $\uparrow$ & \bf R-2 $\uparrow$ & \bf R-L $\uparrow$ & \bf BERTScore $\uparrow$\\
\midrule
\rowcolor[gray]{0.85}
\Block[c]{1-12}{Instruction-following LLMs} \\
\midrule
1 & GPT-5 & 0.67 & 12.26 & 32.04 & 12.50 & 3.79 & 1.06 & 17.88 & 2.01 & 15.29 & 11.25 \\
2 & DeepSeek-R1 & 12.85 & 18.61 & 14.07 & 12.67 & 4.14 & 1.57 & 16.25 & 2.10 & 14.06 & 15.03 \\
3 & Qwen3-235b-Instruct & 15.15 & 19.77 & 13.69 & 11.72 & 3.53 & 1.37 & 15.70 & 1.58 & 13.51 & 12.44 \\
\midrule

\rowcolor[gray]{0.85}
\Block[c]{1-12}{Fine-tuned LLaMA3.1-8b Models} \\
%\multicolumn{12}{c}{\cellcolor{gray}{Fine-tuned LLaMA3.1-8b Models}}\\
\midrule
% 4 & Single Strategy & 24.95	& 28.59	& 12.66	& 16.25	& 6.69 & 2.89 & 17.83 & 17.77 & 5.08 & 27.80 \\
4 & Single Strategy &  25.21 & 28.28 & 12.70 & 16.54 & 6.98 & 3.01 & 22.95 & 4.63 & 18.06 & 18.16\\
\hdashline
5 & All-In-One & 23.61 & 28.63 & 12.50 & 17.04 & 7.19 & 3.06 & 23.33 & 4.74 & 18.27 & 18.17 \\
6 & \quad\quad + Reasoning & 29.72 & 34.92 & 12.37 & \cellcolor{lightred!85}{18.05} & \cellcolor{lightred!85}{8.39} & \cellcolor{lightred!85}{3.98} & 24.96 & \cellcolor{lightred!85}{6.19} & 20.10 & 19.98 \\
7 & \quad\quad + Reasoning + RL & \cellcolor{lightred!85}{29.97}	& \cellcolor{lightred!85}{36.22} & 12.43	& \cellcolor{darkred!85}{18.75}	& \cellcolor{darkred!85}{8.62} & \cellcolor{darkred!85}{4.07} & \cellcolor{darkred!85}{25.68} & \cellcolor{darkred!85}{6.26} & \cellcolor{darkred!85}{20.71} & 20.68 \\
\hdashline
8 & One-by-One & 24.99 & 30.15 & 12.66 & 16.97 & 7.49 & 3.30 & 24.02 & 5.23 & 19.26 & 18.95 \\
9 & \quad\quad + Reasoning & 29.55 & 34.30 & \cellcolor{lightred!85}{12.34} & 17.76 & 8.28 & 3.88 & 24.91 & 6.06 & 20.23 & \cellcolor{lightred!85}{20.72}	\\
10 & \quad\quad + Reasoning + RL & \cellcolor{darkred!85}{33.53} & \cellcolor{darkred!85}{37.97} & \cellcolor{darkred!85}{12.13} & 17.76 & 8.14 & 3.71 & \cellcolor{lightred!85}{25.01} & 5.98 & \cellcolor{lightred!85}{20.38} & \cellcolor{darkred!85}{21.11} \\
\bottomrule
\end{NiceTabular}
}
\caption{Utterance-level evaluation results on the ESC test set. Best scores are shown in \colorbox{darkred!85}{dark red} and second-best are in \colorbox{lightred!85}{light red}. EMR and LR are metrics for strategy prediction while the others are for utterance generation.}
\label{tab:result_utterance}
\end{table*}

\noindent\textbf{Dataset.} We use ESConv and follow the training, validation and testing splitting as in~\citet{liu-etal-2021-esconv}, shown in Table~\ref{tab:stat_esconv}. Each dialogue in ESConv is annotated with a problem type, an emotion type, and a situation description. These annotations are used for self-play dialogue evaluation.

\noindent\textbf{Model Training.} We adopt LLaMA-3.1-8B-Instruct as the backbone and train it using SFT with LoRA and RL with GRPO. For the number of instances collected in SFT and RL, and detailed of hyper-parameter settings in SFT and RL, please refer to Appendix~\ref{apx:training}.  

\noindent\textbf{Baselines.} In ESConv, approximately 82.3\% of supporter utterances contain only a single strategy. Modeling multiple strategies per turn increases generation complexity and may introduce additional noise. To provide a meaningful comparison, we include a \texttt{Single-Strategy} baseline, which assumes that each utterance contains exactly one strategy and therefore predicts only a single strategy-response pair.\footnote{Appendix~\ref{apx:prompt_single_strategy} presents the prompt for the \texttt{Single-Strategy} baseline.} This setting mirrors the conventional ESC formulation and serves as a baseline for evaluating the benefit of our All-in-One and One-by-One approaches\ignore{ that allow one or more strategies per utterance}. In addition, we compare against strong instruction-following LLMs using One-by-One method, including GPT-5, DeepSeek-R1, and Qwen3-235B-Instruct. % For comparison of instruction-following LLMs using the All-in-One method, please refer to Appendix ~\ref{apx:pe_all_in_one}.  

\subsection{Experimental Results: Utterance-level Evaluation} 

\noindent\textbf{Metrics.} Following response-level evaluation, we assess utterance quality using BLEU-1/2/4 (B-1/2/4)~\cite{papineni-etal-2002-bleu}, ROUGE-1/2/L (R-1/2/L)~\cite{lin-2004-rouge}, and BERTScore~\cite{zhang-etal-2019-bertscore}. For strategy prediction, we follow \citet{bai-etal-2025-emotional} and report Exact Match Rate (EMR), which requires an exact match between predicted and reference strategy sequences, and Levenshtein Ratio (LR), which measures similarity between the predicted and reference sequences using a normalized edit distance (Eq.~\ref{equ:lr}), and Average Length Difference (ADL), which examines how the lengths of utterances generated by our models compare to those references.

\subsubsection{Main Results.}
Table~\ref{tab:result_utterance} shows the results on the ESC test set. 

\begin{itemize}[leftmargin=*]
\item All fine-tuned LLaMA3.1-8b models (\#4 - \#10) outperform the instruction-following LLMs (\#1 - \#3). Comparing \#4, \#5 and \#8, we observe that allowing multiple strategies in the All-in-One method reduces EMR compared to the Single-Strategy baseline (23.61 vs. 25.21), likely due to the additional noise introduced when predicting multiple strategies jointly. In contrast, the One-by-One method maintains an EMR close to the baseline (24.99 vs. 25.21), indicating that its iterative formulation effectively mitigates this issue. Notably, despite the minor EMR drop for All-in-One, both All-in-One and One-by-One methods surpass the Single-Strategy baseline, demonstrating that allowing multiple strategies within a single utterance provides meaningful benefits rather than introducing harmful complexity. Moreover, the One-by-One method achieves the strongest performance. This suggests that decomposing multi-strategy generation into sequential prediction steps enables the model to reason more accurately about each strategy and maintain better control over strategy ordering, ultimately leading to higher-quality supportive responses.  
\item By comparing \#5 vs. \#6 and \#8 vs. \#9, we observe that adding cognitive reasoning consistently improves performance across metrics in both All-in-One and One-by-One methods. This demonstrates that explicit reasoning not only helps the model identify the appropriate supportive strategies more accurately, but also guides it toward generating higher-quality responses that better align with the user's emotional needs.
\item Finally, comparing \#6 vs. \#7, we observe that reinforcement learning consistently improves performance for the All-in-One method across all evaluation metrics. In contrast, when comparing \#9 vs. \#10, reinforcement learning leads to improvements on most metrics for the One-by-One method, while yielding slightly lower BLEU scores. Overall, these results indicate that reinforcement learning is effective in enhancing strategy modeling and supportive behavior beyond supervised training, although its impact on surface-level metrics such as BLEU can vary.
\end{itemize}

% Figure~\ref{fig:utterance_case} in Appendix~\ref{apx:dialogue_example} shows examples of utterance-level generation. 
% For performance of instruction-following LLMs using the All-in-One method, please refer to Appendix ~\ref{apx:pe_all_in_one}.

For performance of instruction-following LLMs using the All-in-One method, please refer to Appendix~\ref{apx:pe_all_in_one}. Examples of utterance-level generation are shown in Figure~\ref{fig:utterance_case} in Appendix~\ref{apx:dialogue_example}.

\subsubsection{Discussion}
\noindent\textbf{Performance on Utterances with Single and Multiple Strategies.}\ignore{ To better understand how modeling multiple strategy-response pairs affects different types of utterances, w}We further analyze performance by separating utterances that contain a single strategy (1,937 utterances) from those that contain multiple strategies (452 utterances). This breakdown allows us to examine whether our approaches improve performance on multi-strategy utterances while preserving reasonable performance on the more common single-strategy cases. 

As shown in Table~\ref{tab:detail_utterance}, for utterances with a single strategy, both All-in-One and One-by-One slightly underperform the \texttt{Single-Strategy} baseline, indicating that modeling multiple strategies can introduce noise for these simpler cases. However, incorporating cognitive reasoning substantially improves performance across all metrics, and reinforcement learning further stabilizes and enhances utterance generation. 

In contrast, for utterances with multiple strategies, the \texttt{Single-Strategy} baseline performs poorly, as it cannot model multiple strategies by design. Both All-in-One and One-by-One achieve clear gains in this setting, despite initially low EMR scores. Notably, the addition of cognitive reasoning and reinforcement learning leads to consistent and significant improvements across EMR and generation quality metrics. These results indicate that our approaches are particularly effective for multi-strategy utterances, and that reasoning and reinforcement learning play a crucial role in enabling accurate strategy sequencing and higher-quality supportive responses.

\begin{table}[!t]
\centering
\small
\begin{NiceTabular}{l|llll}
\toprule
\bf Model & \bf EMR & \bf B-4 & \bf R-L & \bf BERTScore \\
\midrule
\rowcolor[gray]{0.85}
\Block[c]{1-5}{Utterances with Single Strategy}\\
%\multicolumn{5}{c}{\cellcolor{gray}{Utterances with Single Strategy}}\\
\midrule
Single Strategy & 31.09 & 3.71 & 19.69 & 19.74\\
\hdashline
All-in-One & 28.76 & 3.34 & 18.72 & 18.87 \\
\quad\quad + Rea. & 35.17 & \cellcolor{lightred!85}{4.41} & 20.67 & 20.62 \\
\quad\quad + Rea. + RL & 34.42 & \cellcolor{darkred!85}{4.47} & \cellcolor{darkred!85}{21.38} & \cellcolor{lightred!85}{21.48} \\
\hdashline
One-by-One & 30.72 & 3.61 & 19.70 & 19.57 \\
\quad\quad + Rea. & 35.46 & 4.26 & 20.88 & 21.56 \\
\quad\quad + Rea. + RL & \cellcolor{darkred!85}{38.24} & 4.05 & \cellcolor{lightred!85}{21.03} & \cellcolor{darkred!85}{21.86} \\
\midrule
% \hline
\rowcolor[gray]{0.85}
\Block[c]{1-5}{Utterances with Multiple Strategies}\\
%\multicolumn{5}{c}{\cellcolor{gray}{Utterances with Multiple Strategies}}\\
% \midrule
\hline
Single Strategy & 0.00 & 1.32 & 14.19 & 13.90 \\
\hdashline
All-in-One & 1.55 & 1.86 & 16.34 & 15.15 \\
\quad\quad + Rea. & 6.35 & 2.20	& 17.64	& 17.21 \\
\quad\quad + Rea. + RL & \cellcolor{lightred!85}{10.88} & \cellcolor{darkred!85}{2.93} & \cellcolor{lightred!85}{18.12} & \cellcolor{lightred!85}{17.86} \\
\hdashline
One-by-One & 0.44 & 2.00 & 17.41 & 16.31 \\
\quad\quad + Rea. & 4.24 & 2.25	& 17.45	& 17.14 \\
\quad\quad + Rea. + RL & \cellcolor{darkred!85}{13.36} & \cellcolor{lightred!85}{2.75} & \cellcolor{darkred!85}{18.17} & \cellcolor{darkred!85}{17.92} \\
\bottomrule
\end{NiceTabular}
\caption{Results on utterances with single and multiple strategies.}
\label{tab:detail_utterance}
\end{table}

% \begin{table}[!h]
% \small
% \centering
% \begin{NiceTabular}{l|l}
% \toprule
% \bf Model & \bf Percentage (\%) \\
% \midrule
% Reference & 18.9 \\
% \midrule
% Single Strategy & 0.0\\
% \midrule
% All-in-One & 2.7 \\
% \quad\quad + Rea. & 6.3 \\
% \quad\quad + Rea. + RL & 8.4 \\
% \midrule
% One-by-One & 1.1 \\
% \quad\quad + Rea. & 3.6 \\
% \quad\quad + Rea. + RL & 7.7 \\
% \bottomrule
% \end{NiceTabular}
% \caption{Percentage (\%) of predicted utterances containing two or more strategies. }
% \label{tab:analysis_utterance}
% \end{table}

\begin{table}[!h]
\small
\centering
\resizebox{\columnwidth}{!}{
\begin{NiceTabular}{l|l||l|l}
\toprule
\bf Model & \bf Per. & \bf Model & \bf Per. \\
\midrule
Reference & 18.9 & Single Strategy & 0.0\\
%\midrule
%Single Strategy & 0.0 \\
\midrule
All-in-One & 2.7 & One-by-One & 1.1 \\
\quad\quad + Rea. & 6.3 & \quad\quad + Rea. & 3.6 \\
\quad\quad + Rea. + RL & 8.4 & \quad\quad + Rea. + RL & 7.7 \\
%\midrule
%One-by-One & 1.1 \\
%\quad\quad + Rea. & 3.6 \\
%\quad\quad + Rea. + RL & 7.7 \\
\bottomrule
\end{NiceTabular}
}
\caption{Percentage (\%) of predicted utterances containing two or more strategies. }
\label{tab:analysis_utterance}
\end{table}

%We further evaluate whether the generated strategies align with the content of the utterance. For example, does {\it Empathy} actually reflect emotional acknowledgment?

\noindent\textbf{Analysis of Utterances with Multiple Strategies.} 
We analyze how often our methods generate supportive utterances containing two or more strategies. As shown in Table~\ref{tab:analysis_utterance}, the ESConv references exhibit a substantial proportion of multi-strategy utterances, with 18.9\% of utterances containing two or more strategies. \ignore{This reflects the natural flexibility of human supporters, who often combine multiple supportive intentions within a single turn.

}In contrast, both All-in-One and One-by-One generate multi-strategy utterances much less frequently in their base settings, with percentages of only 2.7\% and 1.1\%, respectively. This gap is expected, as the training data is dominated by single-strategy instances, which biases models toward simpler generations. Incorporating cognitive reasoning substantially increases the proportion of multi-strategy utterances for both methods. Further applying reinforcement learning leads to additional gains, raising the percentages to 8.4\% for All-in-One and 7.7\% for One-by-One.

\ignore{
\begin{table}[!t]
\small
\centering
\begin{NiceTabular}{l|c|c|c}
\toprule
\bf Model & \bf Base & \bf +Rea. & \bf +Rea. + RL \\
\midrule
Reference & 18.9 & -- & -- \\
Single Strategy & 0.0 & -- & -- \\
\midrule
All-in-One & 2.7 & 6.3 & 8.4 \\
One-by-One & 1.1 & 3.6 & 7.7 \\
\bottomrule
\end{NiceTabular}
\caption{Percentage of predicted utterances containing two or more strategies(all values in \%).}
\label{tab:analysis_utterance}
\end{table}
}

%Taking the model \textit{One-by-One + Reasoning + RL} as example, the most common strategy sequences are ****. 

\noindent\textbf{Effect of Distilling Reasoning from Different LLMs.} 
We analyze how the choice of teacher models influences reasoning quality by comparing the All-in-One method distilled from each individual LLM versus from all four LLMs combined. As shown in Table~\ref{tab:result_reasoning}, distilling reasoning from any single LLM already improves performance compared to training without reasoning, indicating that structured cognitive chains provide valuable supervisory signals. Interesting, no single LLM dominates across all metrics: GPT-5 performs best on BLEU-4, Gemini achieves the highest ROUGE-L, and DeepSeek-R1 yields the strongest BERTScore and EMR. This diversity suggests that different teacher models emphasize complementary aspects of supportive generation. When reasoning chains from all four LLMs are combined, the model attains the most consistent and overall strongest performance across all metrics, highlighting the benefit of aggregating heterogeneous reasoning signals.

\begin{table}[!t]
\small
\centering
\begin{NiceTabular}{l|llll}
\toprule
\bf LLM for Rea. & \bf EMR & \bf B-4 & \bf R-L & \bf BERTScore \\
\midrule
No Reasoning & 23.61 & 3.06 & 18.27 & 18.17\\
\midrule
DeepSeek-R1 & \cellcolor{lightred!85}{28.46} & 3.68 & 19.60 & \cellcolor{lightred!85}{19.32} \\
Qwen3 & 27.25 & 3.50 & 19.45 & 18.92\\
GPT-5 & 26.66 & \cellcolor{lightred!85}{3.72} & 19.84 & 18.99\\
Gemini & 28.05 & 3.54 & \cellcolor{lightred!85}{19.97} & 18.35\\
\midrule 
All & \cellcolor{darkred!85}{29.72} & \cellcolor{darkred!85}{3.98} & \cellcolor{darkred!85}{20.10} & \cellcolor{darkred!85}{19.98}\\
\bottomrule
\end{NiceTabular}
\caption{Performance comparison on the ESC test set when distilling reasoning from different LLMs.}
\label{tab:result_reasoning}
\end{table}

\subsection{Experimental Results: Dialogue-level Evaluation}

\begin{table}[!t]
\centering
\small
%\resizebox{\columnwidth}{!}{
\begin{NiceTabular}{ll|lll}
\toprule
\bf \# &\bf Model & \bf AT$\downarrow$ & \bf AS$\downarrow$ & \bf SR$\uparrow$ \\
\midrule
\rowcolor[gray]{0.85}
\Block[c]{1-5}{Instruction-only LLMs}\\
%\multicolumn{5}{c}{\cellcolor{gray}{Instruction-only LLMs}} \\
\midrule
% 1 & GPT-5 & 10.00 & 1.0960 & 0.00 \\
% 2 & DeepSeek-R1 & 10.00	& 1.0300 & 0.00 \\
% 3 & Qwen3-235b-Instruct & 10.00	& 1.0200 & 0.00 \\
% 4 & Single Strategy & 9.56 & \cellcolor{darkred!85}{1.0000}	& 14.00 \\
% 5 & All-in-One & 9.50 & 1.0140 & 18.00 \\
% 6 & \quad\quad + Rea. &  \cellcolor{lightred!85}{8.80} & 1.0727 & 32.00 \\
% 7 & \quad\quad + Rea. + RL & \cellcolor{lightred!85}{8.80} & 1.1250 & \cellcolor{lightred!85}{34.00} \\
% 8 & One-by-One & 9.72 & \cellcolor{lightred!85}{1.0021} & 16.00 \\
% 9 & \quad\quad + Rea. & 9.08 & 1.0264 & 26.00 \\
% 10 & \quad\quad + Rea. + RL & \cellcolor{darkred!85}{8.46} & 1.2162 & \cellcolor{darkred!85}{40.00} \\
1 & GPT-5 & 10.00 & 10.96 & 0.00 \\
2 & DeepSeek-R1 & 10.00	& 10.30 & 0.00 \\
3 & Qwen3-235b-Instruct & 10.00	& 10.20 & 0.00 \\
\midrule
\rowcolor[gray]{0.85}
\Block[c]{1-5}{Fine-tuned LLaMA3.1-8b Models}\\
%\multicolumn{5}{c}{\cellcolor{gray}{Fine-tuned LLaMA3.1-8b Models}}\\
\midrule
4 & Single Strategy & 9.56 & 9.56 & 13.85 \\
\hdashline
5 & All-in-One & 9.50 & 9.63 & 17.69 \\
6 & \quad\quad + Rea. &  \cellcolor{lightred!85}{8.80} & \cellcolor{lightred!85}{9.44} & 32.31 \\
7 & \quad\quad + Rea. + RL & \cellcolor{lightred!85}{8.80} & 9.90 & \cellcolor{lightred!85}{34.62} \\
\hdashline
8 & One-by-One & 9.72 & 9.74 & 16.15 \\
9 & \quad\quad + Rea. & 9.08 & \cellcolor{darkred!85}{9.32} & 26.15 \\
10 & \quad\quad + Rea. + RL & \cellcolor{darkred!85}{8.46} & 10.29 & \cellcolor{darkred!85}{40.00} \\
\bottomrule
\end{NiceTabular}
%}
\caption{Dialogue-level evaluation results on the ESC test set.}
\label{tab:result_dialogue}
\end{table}

\noindent\textbf{Self-Play Setup.}  We assess dialogue-level performance using a self-play setup following prior work \citep{deng-2024-plug-and-play, he-2025-proactive-dialogues, he-2025-user-tailord-dialogue}. 
\ignore{Recent studies have shown that self-play evaluation can be substantially biased when relying on earlier-generation LLMs (e.g., gpt‑3.5‑turbo), whose lenient criteria often overestimate system performance. In line with observations from \citet{kim-etal-2025-principles} that stronger evaluators provide more reliable judgments, we adopt a more capable model to simulate the seeker and to serve as the reward model.}
In line with observations from \citet{kim-etal-2025-principles} that stronger evaluators provide more reliable judgments, we adopt a more capable model to simulate the seeker and to serve as the reward model.
In this setting, GPT-5 (gpt-5-2025-08-07) simulates the seeker, while our models act as the supporter and generate utterances conditioned on the evolving dialogue history. After each turn, we use GPT-5 as the reward model to evaluate how much progress has been made toward alleviating the seeker's emotional distress. It categorizes the seeker's emotional state into four levels and maps these levels to scalar rewards; 10 sampled evaluations are averaged to produce a continuous final-turn score. A dialogue is deemed successful if this final score exceeds a predefined threshold. If the interaction reaches 10 turns without meeting this criterion, it is counted as a failure. Further prompt details are provided in the Appendix~\ref{apx:prompt_dialogue}. %We evaluate dialogue-level performance using a self-play framework following prior work \citep{deng-2024-plug-and-play, he-2025-proactive-dialogues, he-2025-user-tailord-dialogue}. Two LLMs simulate the supporter and the seeker, and the supporter generates responses according to its predicted dialogue strategies. After each turn, a reward model evaluates the progress toward resolving the seeker's emotional issue. It classifies the seeker's emotional state into four levels and maps them to scalar rewards; multiple sampled judgments are averaged to produce a continuous final-turn reward. A dialogue is considered successful if this reward exceeds a predefined threshold. Dialogues exceeding 10 turns without reaching the goal are treated as failures. Additional implementation details can be found in the Appendix.

\noindent\textbf{Metrics.} Following \citet{deng-2024-plug-and-play}, we report Average Turn (AT) and Success Rate (SR). AT measures goal-completion efficiency by computing the average number of turns required to reach the goal. SR measures goal-completion effectiveness as the proportion of dialogues that achieve the goal within a fixed maximum number of turns. In addition, we introduce Average Strategy (AS), which is defined as the mean number of support strategies used per dialogue. 

\subsubsection{Main Results} 

Table~\ref{tab:result_dialogue} presents the dialogue-level evaluation results. The instruction-only LLMs (\#1$\sim$\#3) fail to resolve the dialogue within the maximum of 10 turns, yielding zero success rate. This highlights the difficulty of emotional support dialogue without task-specific fine-tuning.

For the fine-tuned LLaMA3.1-8B models, both All-in-One and One-by-One (\#5 and \#8) achieve average turn (AT) and average strategy (AS) values comparable to the \texttt{Single-Strategy} baseline (\#4), while yielding noticeably higher success rates (SR). This indicates that allowing multiple strategies per utterance improves the likelihood of successfully resolving the seeker’s emotional issue without increasing dialogue length or strategy usage.

When enhanced with cognitive reasoning and reinforcement learning (\#6$\sim$\#7 and \#9$\sim$\#10), both methods show further improvements. Specifically, these models tend to generate utterances with multiple strategies more effectively, leading to fewer dialogue turns, slightly higher AS scores, and substantially higher success rates. Notably, the One-by-One method with reasoning and RL (\#10) achieves the best overall performance, with the lowest AT and the highest SR. These results suggest that multi-strategy generation, especially when combined with structured reasoning and reinforcement learning, enables more efficient and effective emotional support at the dialogue level.

Figure~\ref{fig:r1_case} $\sim$ \ref{fig:one_by_one_case} in Appendix~\ref{apx:dialogue_example} show dialogue examples.

\subsubsection{Discussion}

\noindent\textbf{Human Evaluation.} Following \citet{zhang-etal-2025-intentionesc}, we conduct human evaluation on complete generated dialogues along four dimensions: \textit{Identification}, \textit{Comforting}, \textit{Suggestion}, and \textit{Overall}. 

Specifically, we randomly sample 50 dialogue annotations from the test set. Based on dialogue annotations, we use GPT‑5 to simulate the seeker's interaction with each of the three systems under evaluation.\ignore{GPT‑5 is employed to extract the seeker's personal information from each dialogue, which is then combined with a associated situation description to simulate the seeker's interaction with each of the three systems under evaluation.} We ask three professional annotators to independently rank the generated dialogues on a three-point scale (1 best, 3 worst) for each evaluation dimension. Finally, we report the average rank across all dialogues. Detained prompts and the annotation guideline are provided in Appendix~\ref{apx:prompt_human_eval} and Appendix~\ref{apx:prompt}.

As shown in Table~\ref{tab:human}, both multi-strategy methods achieve better human rankings than the \texttt{Single-Strategy} baseline across all dimensions. Although these gains are partly influenced by cognitive reasoning and reinforcement learning, the results suggest that flexible multi-strategy utterances are preferred by human annotators.

\begin{table}[!t]
\small
\centering
\resizebox{\columnwidth}{!}{
\begin{NiceTabular}{l|llll}
\toprule
\bf Model & \bf Id. $\downarrow$ & \bf Cm. $\downarrow$ & \bf Sg. $\downarrow$ & \bf Oa. $\downarrow$ \\
\midrule
Single Strategy & 2.18 & 2.24 & 2.40 & 2.34 \\
All-in-One + Rea. + RL. & 1.90 & 1.84 & 1.62 & 1.66 \\
One-by-One + Rea. + RL. & 1.92 & 1.92 & 1.98 & 2.00 \\
\bottomrule
\end{NiceTabular}
}
\caption{Human evaluation results. {\it Id.} denotes Identification, {\it Cm.} denotes Comforting, {\it Sg.} denotes Suggestion, and {\it Oa.} denotes Overall.}
\label{tab:human}
\end{table}

\begin{figure}[t]
\centering
\includegraphics[width=1.0\columnwidth]{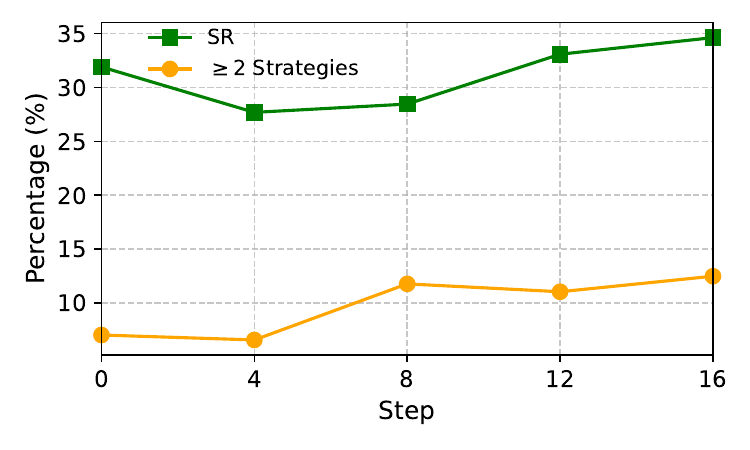}
\caption{Performance comparison across RL training steps.}
\label{fig:encourage_multiple_strategies}
\end{figure}

\noindent\textbf{Reinforcement Learning Curve.} We use the All-in-One method to illustrate the learning curve under reinforcement learning. As shown in Figure~\ref{fig:encourage_multiple_strategies}, during the initial training phase (from the start to step 4), the proportion of multi-strategy utterances slightly decreases, which is accompanied by a drop in dialogue-level SR. After step 4, the model gradually increases its use of multiple strategies, and this trend coincides with the generation of dialogues achieving higher SR. Overall, the learning curve suggests that reinforcement learning eventually encourages more effective multi-strategy usage (from 7.05 at step 0 to 12.50 at step 16), leading to improved dialogue performance (from 32.31 to 34.62).

\ignore{
\noindent\textbf{Analysis of Supportive Utterances with Multiple Strategies.} We find that only about 5.0\% of generated utterances from both methods contain multiple strategies, substantially lower than the 18.9\% in the ESConv test set. Taking the All-in-One method as an example, this gap is likely due to the dominance of single-strategy instances during SFT, which biases the model toward single-strategy generation. 

%To encourage the generation of utterances that include two or more strategies, we modify the RL reward to explicitly favor such cases. Specifically, we add a bonus when the generated output contains more than one strategy. The revised reward is:
To encourage multi-strategy utterances, we modify the RL reward by adding a bonus when more than one strategy is generated:
\begin{equation}
\scriptstyle
r_1\left(y\right) = 
\begin{cases}
\scriptstyle
\text{LR}\left(s_y, s_r\right) & \text{if } r_{\text{fmt}}\left(y\right)=1 \text{\&\&} c\left(y\right)=1, \\
\scriptstyle
\text{LR}\left(s_y, s_r\right) + 0.1 & \text{if } r_{\text{fmt}}\left(y\right)=1 \text{\&\&} c\left(y\right)>1, \\
\scriptstyle
0 & \text{otherwise,}
\end{cases}
\label{equ:new_lr}
\end{equation}
where $c\left(y\right)$ denotes the number of strategies in $y$.
%As shown in Figure~\ref{fig:encourage_multiple_strategies}, the modified reward steadily increases the use of multiple strategies during training. The proportion of utterances containing two or more strategies rises sharply and reaches about 88\% by step 8.

As shown in Figure~\ref{fig:encourage_multiple_strategies}, this reward steadily increases multi-strategy usage, reaching about 88\% by step 8.\footnote{\textcolor{red}{The sharp increase in multi-strategy usage is likely due to the high bonus (0.1 in Eq.~\ref{equ:new_lr}). A smaller value would slow this rise, but we did not explore it further due to API costs.}} At the dialogue level, SR rises alongside increased multi-strategy use and peaks around step 8, but declines when multi-strategy generation becomes overly frequent. This suggests that multi-strategy utterances are beneficial when used appropriately, while excessive use can hinder emotional support effectiveness. 
%At the dialogue level, SR improves in parallel with the increased use of multiple strategies and peaks around step 8. However, as multi-strategy usage becomes overly frequent in later training stages, SR starts to decline. This indicates that while using multiple strategies can enhance dialogue success when applied judiciously, excessive use may negatively affect emotional support quality.
}

\section{Related Work}\label{sec:related_work}
The ESConv dataset proposed by \citet{liu-etal-2021-esconv} has become a widely-used benchmark for ESC research. Most prior work formulates ESC as a combination of support strategy prediction and strategy constrained response generation, treating each supporter turn as being associated with a single strategy. A line of studies leverages external commonsense knowledge to better infer the seeker's emotional state, needs, and persona, thereby improving both strategy selection and response generation \cite{tu-etal-2022-misc,peng-etal-2022-control,cheng-etal-2023-pal,hao-kong-2025-enhancing}. Other approaches explicitly model strategy usage across multiple dialogue turns, capturing global or long-term strategy planning rather than making isolated turn-level decisions \cite{cheng-etal-2022-improving,zhao-etal-2023-transesc}. More recently, some work incorporates explicit reasoning mechanisms, such as chain-of-thought style reasoning, to enhance emotional understanding and supportive planning \cite{chen-etal-2025-socialsim,zhu-etal-2025-care}. In addition, several studies focus on data augmentation by synthetically generating additional training dialogues or utterances to alleviate data sparsity and improve model robustness and generalization~\cite{zheng-etal-2023-augesc,ye-etal-2025-sweetiechat}.

Despite these advances, nearly all existing studies restrict each turn to a single strategy-response pair, leaving the modeling of multiple strategies within a single utterance largely underexplored. To our knowledge, \citet{bai-etal-2025-emotional} is the only work that explicitly investigates multi-strategy generation within a single turn. However, whether allowing multiple strategies in emotional support conversations is beneficial remains unclear. This paper aims to address this open question. 

\section{Conclusion}
In this work, we study emotional support conversation by allowing a supportive utterance to contain one or more strategy-response pairs. To this end, we propose two generation methods, All-in-One and One-by-One, which generate a complete supportive utterance either in a single pass or through iterative strategy-response prediction. Both methods are enhanced with cognitive reasoning and reinforcement learning. Extensive utterance-level and dialogue-level evaluations on ESConv demonstrate that modeling multiple strategies within an utterance is feasible and beneficial, consistently outperforming the \texttt{Single-Strategy} baseline. Our findings highlight the importance of flexible multi-strategy modeling for more natural and effective emotional support systems.

\section*{Limitations}
Despite the improvements demonstrated by our All-in-One and One-by-One methods, several limitations remain. First, the proportion of utterances with multiple strategies generated by our models is still lower than in the ESConv dataset, which may limit the diversity and richness of multi-strategy supportive responses. Second, our evaluation relies on simulated seekers for dialogue-level assessment; while this enables large-scale testing, it may not fully capture the nuances of real human emotional interactions. Third, our study focuses on the ESConv dataset; further evaluation on other emotional support conversation datasets or real-world deployment is needed to assess generalizability.

\section*{Ethical Considerations}
This work investigates emotional support conversation (ESC) systems and proposes methods for generating multi-strategy supportive utterances. Developing models that interact with individuals who may be experiencing distress requires careful ethical consideration. First, while the proposed system is intended to improve empathetic and supportive communication in everyday scenarios, it is not designed or validated for clinical use. The model should not be viewed as a replacement for professional psychological or medical assistance, and its deployment must include safeguards to prevent misuse in high‑risk or crisis situations.

Second, as with all large language model-based systems, our approach carries risks of generating inappropriate, biased, or misleading content. Although we incorporate reasoning mechanisms and reinforcement learning to guide strategy use, these measures alone do not guarantee safety. Additional content moderation and human oversight are necessary for real‑world applications. Developers and practitioners should remain aware of potential biases inherited from training data and ensure that the model’s outputs are used responsibly.

Third, our study adheres to standard privacy and data protection principles. Experiments are conducted exclusively on the ESConv dataset, an established open‑access benchmark that contains no personally identifiable or sensitive user information. No additional user data were collected for this work.

For human evaluation, we recruited three trained annotators from our institution who have prior experience with dialogue assessment. Their participation was voluntary and compensated in accordance with local institutional guidelines, and the payment was considered fair for the time required. No personal or sensitive information about the annotators was collected during the study.

We believe that the ability to compose multiple support strategies within a single utterance may improve the effectiveness of supportive dialogue systems, but it must be deployed with caution to avoid undue reliance on automated systems for emotional or mental health needs.

% Bibliography entries for the entire Anthology, followed by custom entries
%\bibliography{anthology,custom}
% Custom bibliography entries only
\bibliography{custom}

\appendix

\section{Detail of Cognitive Reasoning}
\label{apx:reasoning}

Following \citet{zhu-etal-2025-care}, we adopt a structured cognitive reasoning chain composed of four types of reasoning nodes, each modeling a distinct aspect of the help-seeker's psychological state and the supporter's decision process:
\begin{itemize}[leftmargin=*]
\item \textbf{Context Node} models the external situation and salient emotional cues expressed by the help-seeker, such as academic pressure, interpersonal conflict, or stressful life events. This node provides situational grounding for subsequent reasoning.

\item \textbf{Cognition Node} captures the help-seeker’s internal interpretations, beliefs, or self-assessments regarding the situation, for example, feelings of incompetence or fear of negative evaluation. This component reflects cognitive appraisals that may shape emotional reactions.

\item \textbf{Emotion Node} represents the emotional states that arise from these cognitions, such as anxiety, sadness, or frustration. Explicitly modeling emotions enables the system to tailor supportive responses to the seeker’s affective needs.

\item \textbf{Support Plan Node} specifies the intended supportive strategy or action, such as providing empathy, reframing the situation, or offering actionable suggestions. This node bridges understanding and response generation by translating psychological insight into concrete support.
\end{itemize}

This structured reasoning chain is used to guide both strategy selection and response generation in our models.

\section{Training Details}
\label{apx:training}

Table~\ref{tab:data_stat} shows the number of instances collected for our supervised fine-tuning (SFT) and reinforcement learning (RL).

In the supervised fine-tuning (SFT) stage, we adopt the LLaMA-Factory~\cite{zheng-etal-2024-llamafactory} framework, employing LoRA with a rank of 8 and a scaling factor of 16 for the All-in-One setting, and LoRA with a rank of 16 and a scaling factor of 32 for the One-by-One setting. All SFT experiments are conducted on 4 NVIDIA A100 80GB GPUs, with a per-device batch size of 4 and gradient accumulation over 2 steps. The learning rate is set to 3e-5, and training is performed for 5 epochs. In the reinforcement learning (RL) stage, the SFT models are further optimized using the VERL~\cite{sheng-2024-hybridflow} framework with the GRPO~\cite{shao-etal-2024-deepseekmath} algorithm for 7 epochs on the same hardware setup, with a batch size of 1024, a rollout size of 16, and a learning rate of 1e-6. The KL-penalty coefficient is fixed at 0.01.

Additionally, the data distillation, self-play evaluation, and other processes involving advanced model API calls incurred an approximate total cost of 800 USD.

\begin{table}[!t]
\small
\centering
\begin{NiceTabular}{l|ll}
\toprule
\bf Stage & \bf All-In-One & \bf One-by-One \\
\midrule
SFT w/o reasoning & 10,679 & 12,759\\
SFT w/ reasoning & 42,222 & 51,035 \\
RL & 3,696 & 12,759 \\
\bottomrule
\end{NiceTabular}
\caption{Number of training instances used for SFT w/ and w/o cognitive reasoning, and for RL.}
\label{tab:data_stat}
\end{table}

\section{Performance of Instruction-following LLMs using the All-in-One Method}
\label{apx:pe_all_in_one}

Table \ref{tab:result_utterance_for_all_in_one_pe} presents the utterance-level evaluation results of instruction-following LLMs using the All-in-One method.

\section{Case Study}\label{apx:dialogue_example}

Figure~\ref{fig:utterance_case} illustrates how different systems respond to the same help‑seeker statement in a uniform scenario, enabling a controlled comparison of strategy use and underlying reasoning. By holding the conversational context constant, variations in supportive quality, reasoning depth, and strategy integration become more apparent. Below, we describe the characteristics of each system’s output in turn.

\begin{itemize}[leftmargin=*]
\item \textbf{DeepSeek-R1} This response applies two supportive strategies, offering emotional validation and understanding. However, it does not engage in visible, deliberate reasoning about the situation prior to the reply, which limits its depth and leaves the seeker without a clear course of action.  

\item \textbf{Single-Strategy} Here only one strategy is used, focused on advice-giving. The absence of explicit reflective processing makes the reply appear surface-level, as it jumps directly to a directive without demonstrating consideration of the seeker’s emotional state or interpretive context.  

\item \noindent\textbf{All-in-One + Rea. + RL.} This approach uses two complementary strategies, and it is evident that the reply was preceded by node-based thinking: the supporter considers the external situation, the seeker’s interpretations, the emotional tone, and then formulates a support plan. Integrating these elements in one turn results in a more cohesive blend of empathy and solution, enhancing relational depth and practical utility.  

\item \noindent\textbf{One-by-One + Rea. + RL.} Like \textit{All-in-One + Rea. + RL.}, this response reflects prior node-oriented reasoning, with the supporter first attending to situational and emotional aspects before moving to action-oriented guidance. The sequential format separates emotional alignment from problem-solving, allowing the seeker to process the empathic connection before receiving the proposed plan, which can improve receptivity and trust.  

\end{itemize}

\begin{figure*}[t]
\centering
\includegraphics[width=\textwidth] 
{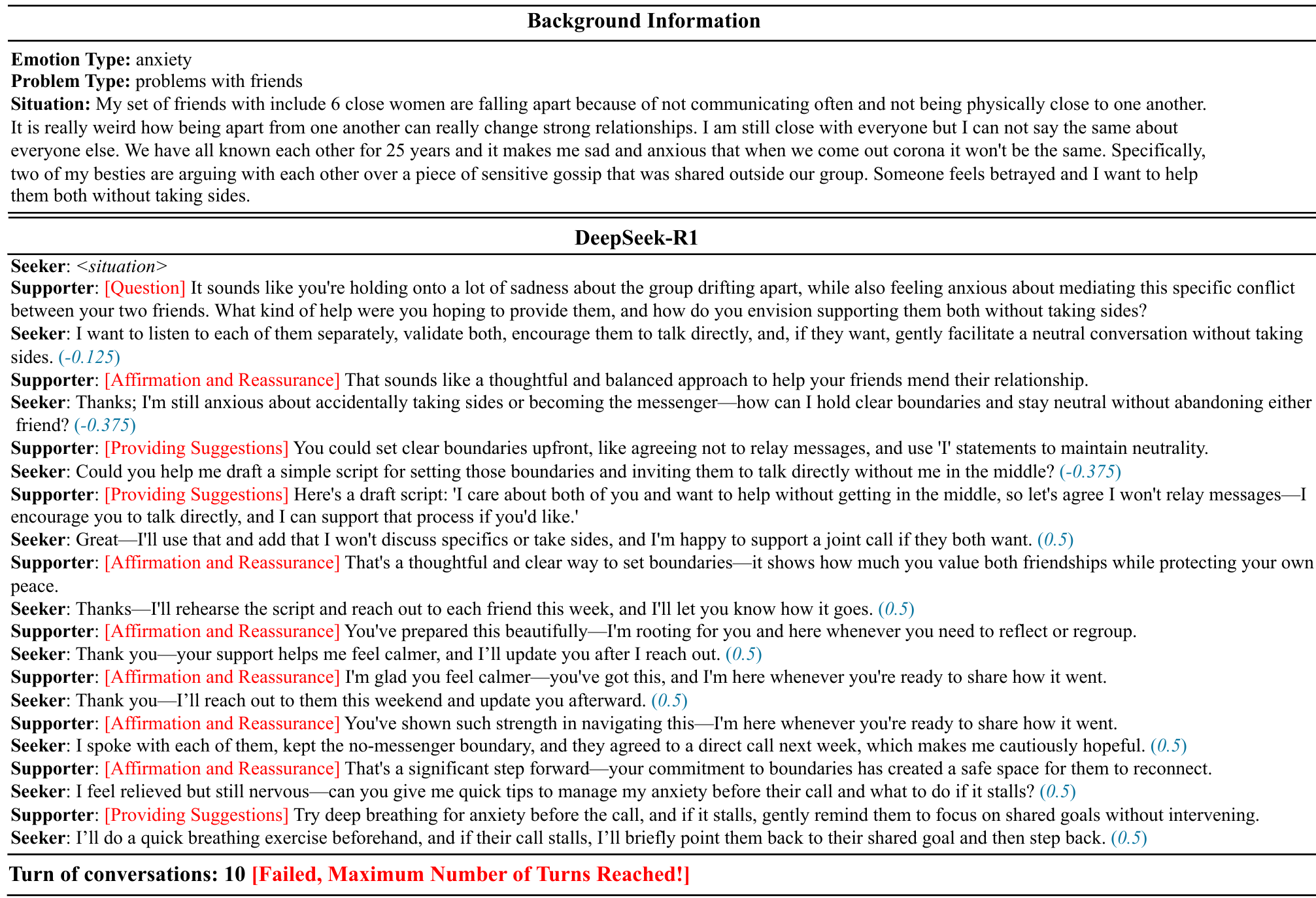}
\caption{Example conversations from DeepSeek-R1.}
\label{fig:r1_case}
\end{figure*}

Figures~\ref{fig:r1_case}, \ref{fig:single_strategy_case}, \ref{fig:all_in_one_case}, and \ref{fig:one_by_one_case} illustrate dialogue examples in an emotional support context, based on a situation where a long-standing friendship group becomes strained during the COVID-19 pandemic and two members fall into conflict over the disclosure of sensitive gossip. The numbers in parentheses indicate the scores assigned by the critic agent to each turn. Notably, the \textit{All‑in‑One + Rea. + RL.} and \textit{One‑by‑One + Rea. + RL.} methods deliver comparable or greater strategic diversity within fewer turns, demonstrating higher efficiency while maintaining empathetic and constructive engagement.

\section{Prompt Details}\label{apx:prompt}

This appendix provides the prompts used across different models and evaluation settings in our experiments.

\subsection{Single-Strategy Baseline}\label{apx:prompt_single_strategy}
Figure~\ref{fig:prompt_baseline} presents the prompt used for the Single-Strategy baseline, which predicts a single support strategy together with its corresponding response for each supporter turn.

\subsection{All-in-One Method}\label{apx:prompt_aio}
For the All-in-One method, Figure~\ref{fig:prompt_all_in_one_pe} shows the prompt without cognitive reasoning, while Figure~\ref{fig:prompt_all_in_one_reasoning} shows the prompt augmented with explicit cognitive reasoning. In addition, Figure~\ref{fig:prompt_all_in_one_reasoning_distill} illustrates the prompt used to distill cognitive reasoning chains from large teacher LLMs.

\subsection{One-by-One Method}\label{apx:prompt_obo}
Figure~\ref{fig:prompt_one_by_one_pe} and Figure~\ref{fig:prompt_one_by_one_reasoning_full} show the prompts for the One-by-One method without and with cognitive reasoning, respectively. Figure~\ref{fig:prompt_one_by_one_reasoning_distill} provides the prompt used for distilling cognitive reasoning from large LLMs in this method.

\subsection{Dialogue-Level Evaluation}\label{apx:prompt_dialogue}
For dialogue-level self-play evaluation, Figure~\ref{fig:prompt_selfplay_user_agent} shows the prompt used to simulate the seeker agent, and Figure~\ref{fig:prompt_selfplay_critic_agent} shows the prompt used by the critic agent to assess dialogue progress and success.  

\subsection{Preparation for Human Evaluation}\label{apx:prompt_human_eval}
Figure~\ref{fig:prompt_profile} and Figure~\ref{fig:prompt_seeker_simulation} show the prompts for profile extraction and seeker simulation, respectively, both adapted from \citet{zhang-etal-2025-intentionesc}.

\section{Guideline of Human Evaluation}\label{apx:guideline}
Figure~\ref{fig:guideline} shows the guideline of human evaluation, adapted from ~\citet{zhang-etal-2025-intentionesc}.

% \begin{table*}[!t]
% \centering
% \small
% \resizebox{\textwidth}{!}{
% \begin{NiceTabular}{ll|ll|llllllll}
% \toprule
% \bf \# & \bf Model & \bf EMR $\uparrow$ & \bf LR $\uparrow$ & \bf ALD $\downarrow$ & \bf B-1 $\uparrow$ & \bf B-2 $\uparrow$ & \bf B-4 $\uparrow$ & \bf R-1 $\uparrow$ & \bf R-2 $\uparrow$ & \bf R-L $\uparrow$ & \bf BERTScore $\uparrow$\\
% \midrule
% 1 & GPT-5 & 1.30 & 19.42 & 34.77 & 13.92 & 4.58	& 1.35 & 18.71 & 2.58 & 16.16 & 12.64 \\
% 2 & DeepSeek-R1 & 2.60 & 21.14 & 76.31 & 10.66 & 3.38 & 0.92 & 16.28 & 1.90 & 14.38 & 3.11 \\
% 3 & Qwen3-235b-Instruct & 1.80 & 18.40 & 84.09 & 9.96 & 3.11 & 0.83 & 15.94 & 1.81 & 14.03	& 2.78 \\
% \bottomrule
% \end{NiceTabular}
% }
% \caption{Utterance-level evaluation results on the ESC test set for instruction-following LLMs using the All-in-One method.} 
% \label{tab:result_utterance_for_all_in_one_pe}
% \end{table*}

% \begin{figure*}[t]
% \centering
% \includegraphics[width=\textwidth] 
% {figure/utterance_case.pdf}
% \caption{Utterance‑level case study on supportive responses from different systems.}
% \label{fig:utterance_case}
% \end{figure*}

% \begin{figure*}[t]
% \centering
% \includegraphics[width=\textwidth] 
% {figure/single_strategy.pdf}
% \caption{Example conversations of Single-Strategy.}
% \label{fig:single_strategy_case}
% \end{figure*}

% \begin{figure*}[t] 
% \centering
% \includegraphics[width=\textwidth] 
% {figure/All-in-One.pdf}
% \caption{Example conversations of \textit{All-in-One + Rea. + RL.}} 
% \label{fig:all_in_one_case}
% \end{figure*}

% \begin{figure*}[t]
% \centering
% \includegraphics[width=\textwidth] 
% {figure/One-by-One.pdf}
% \caption{Example conversations of \textit{One-by-One + Rea. + RL.}}

% \label{fig:one_by_one_case}
% \end{figure*}

% Prompt used for the ESC baseline, which predicts only one support strategy.
\begin{figure*}[!t]
\begin{tcolorbox}[title=Prompt Template, colback=gray!5, colframe=black, fonttitle=\bfseries]
\footnotesize

\# Role \\
You are the supporter in a two-person conversation. The seeker shares their current problem and emotional state. Your task is to apply empathy, build emotional connection, and provide appropriate comfort and support based on the conversation. \\

\#\# Strategy Definitions \\
\text{[Question]: Ask open-ended questions to explore the user’s feelings and situation.} \\
\text{[Restatement or Paraphrasing]: Rephrase what the user said to confirm understanding and show you are listening.} \\
\text{[Reflection of feelings]: Acknowledge and validate the user’s emotions to show empathy.} \\
\text{[Self-disclosure]: Share relevant personal experiences or information when appropriate.} \\
\text{[Affirmation and Reassurance]: Provide comfort and reassurance to reduce the user’s anxiety or distress.} \\
\text{[Providing Suggestions]: Offer practical advice or suggestions to help address the user’s concerns.} \\
\text{[Information]: Provide factual information or explanations relevant to the situation.} \\
\text{[Others]: Responses that do not fit the above categories.} \\ 

\#\# Dialogue Context \\
\{context\} 
\\

\#\# OutputFormat \\
Choose only one strategy that aligns with the dialogue context, and craft your reply accordingly. Strictly follow the JSON format below. 
\begin{verbatim}
{
    "strategy": "your strategy",
    "text": "your response"
}
\end{verbatim}

\end{tcolorbox}
\caption{Prompt used for the ESC baseline, which predicts only one support strategy.}
\label{fig:prompt_baseline}
\end{figure*}

% Prompt used for the All-in-One method, which performs PE.
\begin{figure*}[!t]
\centering
\begin{tcolorbox}[
    title=Prompt Template,
    colback=gray!5,
    colframe=black,
    fonttitle=\bfseries
]
\footnotesize

\# Role \\
You are the supporter in a two-person conversation. The seeker shares their current problem and emotional state. Your task is to apply empathy, build emotional connection, and provide appropriate comfort and support based on the conversation. \\[3pt]

\#\# Strategy Definitions \\
\text{[Question]: Ask open-ended questions to explore the user’s feelings and situation.} \\
\text{[Restatement or Paraphrasing]: Rephrase what the user said to confirm understanding and show you are listening.} \\
\text{[Reflection of feelings]: Acknowledge and validate the user’s emotions to show empathy.} \\
\text{[Self-disclosure]: Share relevant personal experiences or information when appropriate.} \\
\text{[Affirmation and Reassurance]: Provide comfort and reassurance to reduce the user’s anxiety or distress.} \\
\text{[Providing Suggestions]: Offer practical advice or suggestions to help address the user’s concerns.} \\
\text{[Information]: Provide factual information or explanations relevant to the situation.} \\
\text{[Others]: Responses that do not fit the above categories.} \\[3pt]

\#\# Dialogue Context \\
\{context\} \\[3pt]

\#\# OutputFormat \\
You may use one or more strategies in a single reply, but they should be connected and follow the logical flow of the conversation, rather than being independent. Strictly follow the JSON format below. 
\begin{verbatim}
[
    {
        "strategy": "your strategy",
        "text": "your response"
    },
    ...  // if more strategies are needed
]
\end{verbatim}

\end{tcolorbox}

\caption{Prompt used for the All-in-One method, which also performs PE.}  
\label{fig:prompt_all_in_one_pe}
\end{figure*}

% Prompt Template (All-in-One Cognitive Reasoning)
\begin{figure*}[!t]
\centering
\begin{tcolorbox}[
    title=Prompt Template,
    colback=gray!5,
    colframe=black,
    fonttitle=\bfseries
]
\footnotesize

\# Role \\
You are the supporter in a two-person conversation. The seeker shares their current problem and emotional state. Your task is to apply empathy, build emotional connection, and provide appropriate comfort and support based on the conversation. \\

\#\# Strategy Definitions \\
\text{[Question]: Ask open-ended questions to explore the user’s feelings and situation.} \\
\text{[Restatement or Paraphrasing]: Rephrase what the user said to confirm understanding and show you are listening.} \\
\text{[Reflection of feelings]: Acknowledge and validate the user’s emotions to show empathy.} \\
\text{[Self-disclosure]: Share relevant personal experiences or information when appropriate.} \\
\text{[Affirmation and Reassurance]: Provide comfort and reassurance to reduce the user’s anxiety or distress.} \\
\text{[Providing Suggestions]: Offer practical advice or suggestions to help address the user’s concerns.} \\
\text{[Information]: Provide factual information or explanations relevant to the situation.} \\
\text{[Others]: Responses that do not fit the above categories.} \\

\#\# Dialogue Context \\
\{context\} \\

\#\# OutputFormat \\
Present your complete thought process step by step within the <think> tags, following the structured format below. After that, provide your strategy and response text within the <answer> tags. \\

<think> \\
\text{[Context]}: Empathic snapshot of the external situation and salient emotional cues, mainly derived from Last Message, providing situational grounding for understanding.\\
\text{[Cognition]}: Internal interpretations, beliefs, or self-assessments that connect the situation to perceived emotional or practical needs, inferred from Last Message. \\
\text{[Emotion]}: Emotional states that may emerge from these cognitions, particularly if supportive reply is absent; derived from the situation and cognition in Last Message. \\
\text{[Support Plan]}: Identify the intended supportive strategy, and briefly explain its sequence and purpose in relation to this reply’s role. \\
</think>

<answer> 
\begin{verbatim}
[
    {
        "strategy": "your strategy",
        "text": "your response"
    },
    ...  // if using extra strategies
]
\end{verbatim}
</answer>

\end{tcolorbox}

\caption{Prompt used for the All-in-One method, which incorporates explicit cognitive reasoning.}
\label{fig:prompt_all_in_one_reasoning}
\end{figure*}

% Prompt Template (All-in-One Cognitive Reasoning Distillation)
\begin{figure*}[!t]
\centering
\begin{tcolorbox}[
    title=Prompt Template,
    colback=gray!5,
    colframe=black,
    fonttitle=\bfseries
]
\footnotesize

You are a psychological support assistant tasked with explaining a supporter’s decision-making process. \\
Given: \\
1. The ongoing dialogue context (focus on the most recent seeker message). \\
2. The supporter’s final reply, with the strategy tag(s) used, listed in execution order. \\

Your goal: Generate a concise, logically consistent reasoning sequence showing how the supporter arrived at the reply. Use exactly four nodes: Context, Cognition, Emotion, and Support Plan. \\

\#\# Node definitions \\
- Context: One short sentence giving an empathic snapshot of the external situation and key emotional cues from the latest message. \\
- Cognition: One short sentence stating the seeker’s inferred interpretations or beliefs that connect the situation to their needs. \\
- Emotion: One short sentence describing the probable emotional state emerging from these cognitions. \\
- Support Plan: Write all strategies sequentially as Step 1, Step 2, Step 3..., each step including: \\
\quad * The exact strategy tag (copy exactly as given, preserve order). \\
\quad * Add a brief note (\(\leq 15\) words) on how it shaped specific aspects of the final reply’s tone, content, vocabulary, or structure. \\
\quad * Combine all steps into a single string, separated by periods. \\

\#\# Dialogue Context \\
\{context\} \\

\#\# Tagged Supporter Reply \\
\{response\} \\

\#\# Output Format Requirements \\
Return ONLY the reasoning process in valid JSON format: 
\begin{verbatim}
{
    "Context": "One short sentence (max 25 words), starting with 'The seeker ...'",
    "Cognition": "One short sentence (max 25 words), starting with 'The seeker ...'",
    "Emotion": "One short sentence (max 25 words), starting with 'The seeker ...'",
    "Support Plan": "Step 1: Using [StrategyTag1] to <short effect>. 
                 Step 2: Using [StrategyTag2] to <short effect dependent on Step 1>. 
                 Step 3: ..."
}
\end{verbatim}

\#\# Rules \\
1. Focus mainly on the latest seeker message; use earlier context only if directly relevant to their state. \\
2. Copy the strategy tags exactly as shown in \#\# Tagged Supporter Reply — keep order and spelling consistent. \\
3. If a node’s information cannot be inferred logically, write "N/A". \\
4. Each node must be concise — Context/Cognition/Emotion \(\leq 25\) words; Support Plan \(\leq 20\) words each. \\
5. Maintain an objective, third-person analytical tone for Context, Cognition, and Emotion. \\
6. The Support Plan node must clearly reflect the sequence and dependency between strategies, link each strategy directly to concrete reply elements (tone, choice of words, specific content, and structure). \\
7. Output only the JSON structure — no comments, explanations, or additional text outside. \\

\end{tcolorbox}

\caption{Prompt used for the All-in-One method, which distills cognitive reasoning.}
\label{fig:prompt_all_in_one_reasoning_distill}
\end{figure*}

% Prompt Template (One-by-One PE)
\begin{figure*}[!t]
\centering
\begin{tcolorbox}[
    title=Prompt Template,
    colback=gray!5,
    colframe=black,
    fonttitle=\bfseries,
]
\footnotesize

\# Role \\
You are the supporter in a two-person conversation. The seeker shares their current problem and emotional state. Your task is to apply empathy, build emotional connection, and provide appropriate comfort and support based on the conversation. Additionally, determine whether you should continue responding, for example by applying more strategies and offering further replies, or pause and allow the seeker to continue. \\

\#\# Strategy Definitions \\
\text{[Question]: Ask open-ended questions to explore the user’s feelings and situation.} \\
\text{[Restatement or Paraphrasing]: Rephrase what the user said to confirm understanding and show you are listening.} \\
\text{[Reflection of feelings]: Acknowledge and validate the user’s emotions to show empathy.} \\
\text{[Self-disclosure]: Share relevant personal experiences or information when appropriate.} \\
\text{[Affirmation and Reassurance]: Provide comfort and reassurance to reduce the user’s anxiety or distress.} \\
\text{[Providing Suggestions]: Offer practical advice or suggestions to help address the user’s concerns.} \\
\text{[Information]: Provide factual information or explanations relevant to the situation.} \\
\text{[Others]: Responses that do not fit the above categories.} \\

\#\# Dialogue Context \\
\{context\} \\

\#\# OutputFormat \\
Choose only one strategy that aligns with the dialogue context, and craft your reply accordingly. Ensure you also include the continue\_reply field with a boolean value indicating whether you wish to keep replying or let the seeker respond next. Strictly follow the JSON format below. \\

\begin{verbatim}
{
    "strategy": "your strategy",
    "text": "your response",
    "continue_reply": true|false
}
\end{verbatim}

\end{tcolorbox}

\caption{Prompt used for the One-by-One method, which also applies PE.}
\label{fig:prompt_one_by_one_pe}
\end{figure*}

% Prompt Template (One-by-One with Explicit Cognitive Reasoning)
\begin{figure*}[!t]
\centering
\begin{tcolorbox}[
    title=Prompt Template,
    colback=gray!5,
    colframe=black,
    fonttitle=\bfseries
]
\footnotesize

\# Role \\
You are the supporter in a two-person conversation. The seeker shares their current problem and emotional state. Your task is to apply empathy, build emotional connection, and provide appropriate comfort and support based on the conversation. Additionally, determine whether you should continue responding, for example by applying more strategies and offering further replies, or pause and allow the seeker to continue. \\

\#\# Strategy Definitions \\
\text{[Question]: Ask open-ended questions to explore the user’s feelings and situation.} \\
\text{[Restatement or Paraphrasing]: Rephrase what the user said to confirm understanding and show you are listening.} \\
\text{[Reflection of feelings]: Acknowledge and validate the user’s emotions to show empathy.} \\
\text{[Self-disclosure]: Share relevant personal experiences or information when appropriate.} \\
\text{[Affirmation and Reassurance]: Provide comfort and reassurance to reduce the user’s anxiety or distress.} \\
\text{[Providing Suggestions]: Offer practical advice or suggestions to help address the user’s concerns.} \\
\text{[Information]: Provide factual information or explanations relevant to the situation.} \\
\text{[Others]: Responses that do not fit the above categories.} \\

\#\# Dialogue Context \\
\{context\} \\

\#\# OutputFormat \\
Present your complete thought process step by step within the <think> tags, following the structured format below. After that, provide your strategy and response text within the <answer> tags. \\

<think> \\
\text{[Context]}: Empathic snapshot of the external situation and salient emotional cues, mainly derived from Last Message, providing situational grounding for understanding.\\
\text{[Cognition]}: Internal interpretations, beliefs, or self-assessments that connect the situation to perceived emotional or practical needs, inferred from Last Message. \\
\text{[Emotion]}: Emotional states that may emerge from these cognitions, particularly if supportive reply is absent; derived from the situation and cognition in Last Message. \\
\text{[Support Plan]}: Identify the intended supportive strategy, and briefly explain its purpose in relation to this reply’s role. \\
</think>

<answer>
\begin{verbatim}
{
    "strategy": "your strategy",
    "text": "your response",
    "continue_reply": true|false
}
\end{verbatim}
</answer>

\end{tcolorbox}

\caption{Prompt used for the One-by-One method, which incorporates explicit cognitive reasoning.}
\label{fig:prompt_one_by_one_reasoning_full}
\end{figure*}

% Prompt Template (One-by-One Cognitive Reasoning Distillation)
\begin{figure*}[!t]
\centering
\begin{tcolorbox}[
    title=Prompt Template,
    colback=gray!5,
    colframe=black,
    fonttitle=\bfseries
]
\footnotesize

\# Role \\
You are a psychological support assistant tasked with explaining a supporter’s decision-making process. \\ 
Given: \\ 
1. The ongoing dialogue context (focus on the most recent seeker message). \\
2. The supporter’s final reply, with the strategy tag used. \\

Your goal: Generate a concise, logically consistent reasoning sequence showing how the supporter arrived at the reply. Use exactly four nodes: Context, Cognition, Emotion, and Support Plan. \\

\#\# Node definitions \\
- Context: One short sentence giving an empathic snapshot of the external situation and key emotional cues from the latest message. \\
- Cognition: One short sentence stating the seeker’s inferred interpretations or beliefs that connect the situation to their needs. \\
- Emotion: One short sentence describing the probable emotional state emerging from these cognitions. \\
- Support Plan: Including the exact strategy tag (copy exactly as given) and a brief note (\(\leq 15\) words) describing how it shaped tone, content, vocabulary, or structure. \\

\#\# Dialogue Context \\
\{context\} \\

\#\# Tagged Supporter Reply \\
\{response\} \\

\#\# Output Format Requirements \\
Return ONLY the reasoning process in valid JSON format: \\

\begin{verbatim}
{
    "Context": "One short sentence (max 25 words), starting with 'The seeker ...'",
    "Cognition": "One short sentence (max 25 words), starting with 'The seeker ...'",
    "Emotion": "One short sentence (max 25 words), starting with 'The seeker ...'",
    "Support Plan": "Using [StrategyTag] to <short effect>."
}
\end{verbatim}

\#\# Rules \\
1. Focus mainly on the latest seeker message; use earlier context only if directly relevant to their state. \\
2. Copy the strategy tag exactly as shown in \#\# Tagged Supporter Reply — keep spelling consistent. \\
3. If a node’s information cannot be inferred logically, write "N/A". \\
4. Each node must be concise — Context/Cognition/Emotion \(\leq 25\) words; Support Plan \(\leq 25\) words. \\
5. Maintain an objective, third-person analytical tone for Context, Cognition, and Emotion. \\
6. The Support Plan node should link the strategy directly to concrete reply elements (tone, choice of words, specific content, and structure). \\
7. Output only the JSON structure — no comments, explanations, or additional text outside. \\

\end{tcolorbox}

\caption{Prompt used for the One-by-One method, which distills cognitive reasoning.}
\label{fig:prompt_one_by_one_reasoning_distill}
\end{figure*}

% Prompt used for the self-play setting, which represents the user agent.)
\begin{figure*}[!t]
\centering
\begin{tcolorbox}[
    title=Prompt Template,
    colback=gray!5,
    colframe=black,
    fonttitle=\bfseries
]
\footnotesize

\textbf{System:} \\
Now enter the role-playing mode. In the following conversation, you will play as a patient in a counselling conversation with a therapist. \\

\textbf{User:} \\
You are the patient who is looking for help from the therapist, because you have the emotional issue about \{emotion\_type\} regarding \{problem\_type\}. Please reply with only one short and succinct sentence. Now tell me your issue. \\

\textbf{Assistant:} \\
\{situation\} \\

\end{tcolorbox}

\caption{Prompt used for the self-play setting, which represents the user agent.}
\label{fig:prompt_selfplay_user_agent}
\end{figure*}

% Prompt Template (Self-Play Critic Agent)
\begin{figure*}[!t]
\centering
\begin{tcolorbox}[
    title=Prompt Template,
    colback=gray!5,
    colframe=black,
    fonttitle=\bfseries
]
\small

\textbf{System:} \\
Given a conversation between a Therapist and a Patient, please assess whether the Patient's emotional issue has been solved after the conversation. \\

\textbf{User:} \\
You can only reply with one of the following sentences: \\
A. No, the Patient feels worse. \\
B. No, the Patient feels the same. \\
C. No, but the Patient feels better. \\
D. Yes, the Patient's issue has been solved. \\

If you believe that the patient’s problem has been resolved or the patient has realized how to solve the problem, please choose D. \\
If you believe that the patient’s problem has not been fully resolved, but the emotional issue has been somewhat alleviated compared to the last conversation turn, choose C. \\
If you believe that the patient’s emotional state has worsened compared to the last conversation turn, choose A. \\
Otherwise, if the patient’s emotional state remains unchanged, choose B. \\

The following is a conversation about \{emotion\_type\} regarding \{emotion\_type\}: \\
\{conversation\} \\

Question: Has the Patient's issue been solved? \\
Answer: \\
\end{tcolorbox}

\caption{Prompt used for the self-play setting, which represents the critic agent.}
\label{fig:prompt_selfplay_critic_agent}
\end{figure*}

\begin{figure*}[!t]
\centering
\begin{tcolorbox}[
    title=Prompt Template,
    colback=gray!5,
    colframe=black,
    fonttitle=\bfseries
]
\small

Please extract and summarize the Seeker’s personal information from the emotional support conversation between the Seeker and Supporter. Summarize the Seeker’s information objectively in a paragraph. \\
Make sure to fully capture the information provided by the Seeker, without making subjective assumptions. \\

Conversation: \\
\{dialog\} \\

Seeker’s personal\_summary: \\
<to be generated> \\

\end{tcolorbox}

\caption{Prompt used for profile extraction.}
\label{fig:prompt_profile}
\end{figure*}

\begin{figure*}[!t]
\centering
\begin{tcolorbox}[
    title=Prompt Template,
    colback=gray!5,
    colframe=black,
    fonttitle=\bfseries
]
\small

You are currently experiencing \{situation\} and engaging in a conversation for emotional support with a supporter. Below is your personal background: \\

<Personal Summary>: \\
\{personal\_summary\} \\

As a seeker, your task is based on the dialogue history and your current emotional state, to respond naturally while incorporating your genuine emotions. \\

<Notion>: \\
1. If at any point you feel you’ve received enough support or are feeling overwhelmed or exhausted, you may end the conversation by saying, “Thanks. Please stop the conversation now.” \\
2. Avoid sharing too many personal information or speak more than 3 sentences upfront or in a single message. \\

<Dialogue History>: \\
\{dialogue\_history\} \\

\end{tcolorbox}

\caption{Prompt used for seeker simulation.}
\label{fig:prompt_seeker_simulation}
\end{figure*}

\begin{figure*}[!t]
\begin{tcolorbox}[title=Guideline of Human Evaluation, colback=gray!5, colframe=black, fonttitle=\bfseries]
\footnotesize

\# Evaluation Objectives \\
The objective of this evaluation is to evaluate the emotional support performance of three Supporters in entire conversations with the same Seeker. \\
Note: The samples evaluated may contain negative or adverse content. Evaluators must maintain an objective and neutral attitude during evaluation. \\

\# Sample Description \\
Each evaluation sample includes records of emotional support conversations between the same Seeker and three different Supporter. \\

\# Evaluation Dimensions \\
\#\# 1. Identification: Evaluate whether the supporter effectively guides the seeker to deeply explore their own issues and whether they help the seeker view the problem from new perspectives. \\
• Depth of Problem Exploration: Does the Supporter help the Seeker uncover the root causes of their issues through questions or guidance? \\
• Self-Understanding Guidance: Does the Supporter encourage the Seeker to reflect on their self-awareness of the problem and enhance their understanding of their own emotions and situation? \\
• Perspective Expansion: Does the Supporter provide new perspectives, helping the Seeker view their problem or predicament from different angles? \\

\#\# 2. Comforting: Evaluate whether supporters are emotionally capable of effectively comforting seekers and alleviating their negative emotions. \\
• Empathy Display: Does the Supporter show understanding and resonance with the Seeker’s emotions? \\
• Emotional Relief Effectiveness: Does the comfort provided help alleviate the Seeker’s negative emotions? \\
• Tone and Diction of Emotional Support: Is the language used by the Supporter warm, considerate, and calming, making the Seeker feel cared for? \\

\#\# 3. Suggestion: Evaluate whether the suggestions provided by supporters are targeted, feasible, and practically helpful. \\
• Targetedness of Suggestions: Are the suggestions clearly targeted at the Seeker’s specific issues? \\
• Actionability: Are the suggestions feasible, and can the Seeker easily implement them? \\
• Practicality and Effect: Based on the Seeker’s feedback, do the suggestions have a practical impact on the Seeker’s issues? \\

\#\# 4. Overall: Evaluate the overall performance of the supporter by considering problem identification, comforting skills, and the effectiveness of the suggestions provided, ultimately determining whether a good emotional support experience is delivered. \\
• Comprehensive Performance: Does the Supporter perform well across all dimensions, making the Seeker feel understood, supported, and helped? \\
• Overall Satisfaction: From the Seeker’s perspective, did the emotional support meet expectations, and was there an improvement in their emotions? \\

\# Evaluation Steps \\
1. Carefully read the conversations to fully understand the Seeker’s emotional state and support needs. \\
2. Read the three conversations and assess and rank them based on the dimensions of “Identification,” “Comforting,” “Suggestion,” and “Overall.” \\
3. Complete the ranking table, assigning a rank from 1 to 3 for each conversation, where 1 is the best and 3 is the worst. \\

\end{tcolorbox}
\caption{Guideline of human evaluation.}
\label{fig:guideline}
\end{figure*}

\end{document}